\documentclass[acmtog,nonacm]{acmart}

\usepackage{xcolor}
\usepackage{colortbl}

\newcommand{\bestcell}[1]{\cellcolor{green!30!yellow!30}\textbf{#1}}
\newcommand{\secondcell}[1]{\cellcolor{yellow!30}#1}

\AtBeginDocument{%
  \providecommand\BibTeX{{%
    \normalfont B\kern-0.5em{\scshape i\kern-0.25em b}\kern-0.8em\TeX}}}

\begin{document}

\title{InfiniSplat: Implicit Gaussian Decoding for Large-Baseline Monocular View Synthesis}

\author{Jiawei Wang}
\authornote{These authors contributed equally to this work.}
\email{jiawei-wang@zju.edu.cn}
\affiliation{%
  \institution{State Key Laboratory of CAD\&CG, Zhejiang University}
  \country{China}}

\author{Hao Yu}
\authornotemark[1]
\email{ritianyu@zju.edu.cn}
\affiliation{%
  \institution{Zhejiang University}
  \country{China}}

\author{Yongzhen Hu}
\email{22451272@zju.edu.cn}
\affiliation{%
  \institution{Zhejiang University}
  \country{China}}

\author{Xinyi Yang}
\email{3240105610@zju.edu.cn}
\affiliation{%
  \institution{Zhejiang University}
  \country{China}}

\author{Tao Ni}
\email{nitao@udeer.ai}
\affiliation{%
  \institution{Udeer.ai}
  \country{China}}

\author{Xin Zhan}
\email{zhanxin@udeer.ai}
\affiliation{%
  \institution{Udeer.ai}
  \country{China}}

\author{Junbo Chen}
\authornote{Corresponding authors.}
\email{junbo@udeer.ai}
\affiliation{%
  \institution{Udeer.ai}
  \country{China}}

\author{Xiaowei Zhou}
\email{xwzhou@zju.edu.cn}
\affiliation{%
  \institution{State Key Laboratory of CAD\&CG, Zhejiang University}
  \country{China}}

\author{Ruizhen Hu}
\email{ruizhen.hu@gmail.com}
\affiliation{%
  \institution{Shenzhen University}
  \country{China}}

\author{Sida Peng}
\authornotemark[2]
\email{pengsida@zju.edu.cn}
\affiliation{%
  \institution{Zhejiang University}
  \country{China}}

\setlength{\skip\footinsauthorsaddresses}{5pt}

\begin{teaserfigure}
  \centering
  \includegraphics[width=0.99\linewidth]{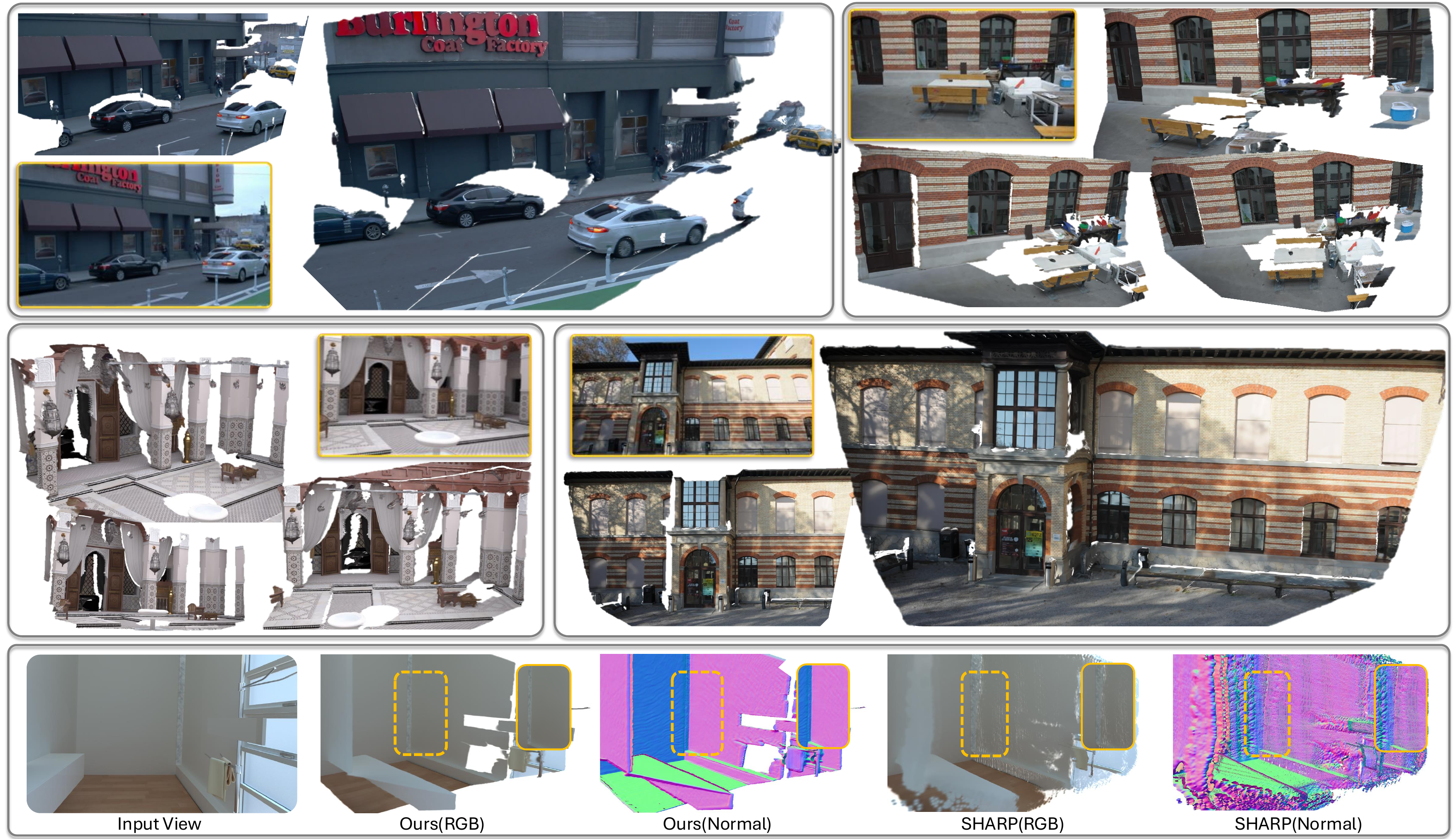}
  \vspace{-3mm}
  \caption{InfiniSplat generates a surface-aligned 3D scene representation from a single image, enabling high-quality novel view synthesis across diverse scenes. Compared with SHARP~\cite{mescheder2025sharp}, InfiniSplat maintains more surface-aligned structures and produces more plausible renderings.}
  \label{fig:teaser}
\end{teaserfigure}

\begin{abstract}
Single-image feed-forward 3D Gaussian Splatting (3DGS) aims to directly generate a renderable 3D scene representation from one input image, avoiding the cost of multi-view capture and per-scene optimization. 
However, existing methods are often constrained by a pixel-aligned representation, where Gaussians are predicted from fixed image-grid locations.
Such pixel-aligned primitives can produce promising nearby-view renderings, but they remain weakly coupled to underlying scene surfaces and struggle to preserve coherent structures under large viewpoint shifts.
We present \textbf{InfiniSplat}, a feed-forward single-image 3DGS framework that moves from a pixel-aligned representation toward a surface-aligned representation.
InfiniSplat constructs this representation by first using geometry-guided sampling to place 2D supports according to depth-induced local surface structure, and then applying a query-conditioned implicit decoder to predict Gaussian attributes from the image features queried at these supports.
By grounding support locations in geometry while decoupling Gaussian prediction from fixed pixel centers, InfiniSplat produces Gaussian layouts that better follow scene surfaces and reduce scattered primitives caused by grid discretization.
Across multiple cross-dataset NVS evaluations, InfiniSplat achieves state-of-the-art performance compared with single-image feed-forward baselines, and demonstrates zero-shot generalization from Hypersim indoor synthetic training to complex open-world scenes.
\textbf{Project page:} \url{https://zju3dv.github.io/InfiniSplat}.
\end{abstract}



\maketitle

\section{Introduction}

Recent advances in neural rendering~\cite{mildenhall2021nerf} and 3D Gaussian Splatting (3DGS)~\cite{kerbl20233d} have substantially improved the rendering efficiency and visual quality of novel view synthesis (NVS).
However, many of the most compelling results still depend on multi-view inputs or require time-consuming per-scene optimization.
In contrast, we focus on a more constrained yet practically attractive setting: quickly generating a renderable 3D scene representation from a single photograph.
This setting can support interactive spatial photo browsing, natural head motion in AR/VR, and spatial content display on handheld devices.
An ideal system in this setting should not only generate a 3D representation in one feed-forward pass, but also maintain stable scene structure as the viewpoint changes, rather than appearing plausible only under tiny pose perturbations.

Existing single-image feed-forward 3DGS methods have made important progress toward predicting explicit Gaussian representations from one image in a single pass~\cite{szymanowicz2025flash3d, mescheder2025sharp}.
However, impressive renderings near the input view do not necessarily imply a Gaussian layout that behaves as a coherent 3D scene under large viewpoint changes.
A central limitation is their pixel-aligned representation: Gaussian primitives are generated from fixed image-grid locations, so they can behave more like locally expanded splats than a surface-aligned scene layout.
When the target camera moves farther from the source view, especially under lateral large-baseline target views, renderings that initially appear high-quality can reveal tearing, surface breaks, loose structure, and severe geometric distortion.
Representative methods such as SHARP achieve strong nearby-view sharpness, but their Gaussian layouts can still struggle to preserve surface-aligned structures under large viewpoint changes.

We identify the pixel-aligned representation as a key bottleneck in current feed-forward single-image Gaussian reconstruction methods.
Our core insight is that single-image 3DGS should move from a pixel-aligned representation toward a surface-aligned representation.
This transition requires two conditions: Gaussian supports should follow depth-induced local surface structures, and Gaussian attribute prediction should not be tied to fixed pixel centers.
Together, these conditions allow neighboring primitives to assemble into coherent surfaces rather than remain independent local splats.

Based on this insight, we propose InfiniSplat, a feed-forward single-image 3DGS framework for surface-aligned Gaussian generation.
InfiniSplat realizes this representation by coupling geometry-guided support sampling with query-conditioned implicit Gaussian decoding.
Geometry-guided sampling uses pretrained monocular geometric priors to place 2D supports according to local surface structure, thereby moving Gaussian supports away from the regular image grid.
The implicit decoder then predicts Gaussian attributes on these sampled supports, allowing a shared decoder to operate on support distributions with different densities and spatial arrangements.
Together, geometry-guided sampling defines surface-aligned support locations, while implicit decoding turns those flexible supports into Gaussian primitives.

Concretely, InfiniSplat employs a dual-branch encoder to build an expressive image feature space, where a DINO branch provides semantic representations and a CNN branch captures local texture cues.
At each sampled support, the decoder queries the DINO and CNN features and predicts Gaussian attributes from the fused support-conditioned descriptor.
Since Gaussian prediction is conditioned on geometry-guided supports rather than tied to the regular pixel grid, InfiniSplat can allocate more primitives to geometrically demanding regions while maintaining consistency among neighboring Gaussians.
This produces more stable structures under large viewpoint changes and reduces scattered layouts caused by grid discretization (see Fig.~\ref{fig:teaser}).

We evaluate InfiniSplat across multiple NVS datasets in a cross-dataset setting.
InfiniSplat achieves state-of-the-art performance across multiple datasets and major image-quality metrics.
These results show that InfiniSplat moves single-image 3DGS beyond view-local splat expansion: by building a surface-aligned Gaussian representation through geometry-guided supports and implicit Gaussian decoding, it maintains more coherent structures under large-baseline viewpoint changes.
In summary, our contributions are as follows:\nopagebreak
\begin{list}{\labelitemi}{%
  \setlength{\leftmargin}{1.5em}
  \setlength{\labelwidth}{1em}
  \setlength{\labelsep}{0.5em}
  \setlength{\topsep}{0.45em}
  \setlength{\itemsep}{0.2em}
  \setlength{\parsep}{0pt}
  \setlength{\parskip}{0pt}
}
  \item \textbf{A surface-aligned Gaussian representation for\\ robust large-baseline NVS.}
  We introduce a geometry-guided support sampling strategy that instantiates this representation by placing 2D supports according to depth-induced local surface structures rather than fixed image-grid locations.
  By aligning support locations with geometric priors, the predicted primitives can better assemble into coherent surfaces, leading to more stable rendering results under large-baseline viewpoint changes.

  \item \textbf{Implicit Gaussian decoding for single-image 3DGS.}
  We formulate Gaussian attribute prediction as query-conditioned implicit decoding over sampled supports and queried image features.
  This decoder turns geometry-guided supports into Gaussian primitives and allows a shared prediction function to operate on support sets with different densities and spatial arrangements.
\end{list}

\section{Related Work}

\subsection{Single-Image Novel View Synthesis}

Single-image novel view synthesis (NVS) aims to synthesize novel views of a scene from a single input image, where the underlying 3D structure is severely under-constrained.
Early methods typically rely on depth-based warping, soft layering, layered depth images, or multiplane image representations to synthesize nearby views~\cite{zhou2016view,wiles2020synsin,jampani2021slide,zhou2018stereo,tucker2020single,han2022single,khan2023tiled}.
These representations provide useful geometric inductive biases, but they are often limited to moderate viewpoint changes and do not directly produce an explicit 3D representation suitable for real-time rendering.

Recent diffusion-based methods and large transformer-based models further improve scene-level generation and large-baseline view synthesis~\cite{hong2024lrm,jin2024lvsm,watson2022novel,liu2023zero,kant2023invs,gu2023nerfdiff,sargent2024zeronvs,gao2024cat3d,yu2025viewcrafter,zhang2025high,zhou2025stable,szymanowicz2025bolt3d,liang2025wonderland,yu2025wonderworld,ren2025gen3c,szymanowicz2026lagernvs}.
However, many of these methods synthesize images directly or rely on iterative generation, and therefore do not naturally provide a compact, explicit, and real-time renderable scene representation.

To obtain an explicit 3D representation, mesh-based single-image-to-3D methods reconstruct geometry and texture from a single image~\cite{liu2023one,long2024wonder3d,wang2024crm,xu2024instantmesh,wu2024unique3d,tochilkin2024triposr,boss2025sf3d,li2024craftsman3d,xiang2025structured,zhao2025hunyuan3d}.
However, these methods primarily focus on object-centric 3D asset reconstruction or generative 3D creation and are not designed for real-scene NVS.
Meanwhile, existing mesh-based scene-level reconstruction methods~\cite{hu2021worldsheet,zhao2025depr,zhang2026pixarmesh} either remain limited in fine-grained geometric and texture fidelity or do not prioritize photorealistic rendering for view synthesis.

More directly related to our work are feed-forward 3DGS methods for single-image NVS, such as Splatter Image~\cite{szymanowicz2024splatter}, Flash3D~\cite{szymanowicz2025flash3d}, SHARP~\cite{mescheder2025sharp}, and ADGaussian~\cite{song2025adgaussian}.
These methods predict explicit 3D Gaussians in a single forward pass for efficient novel-view rendering, but mostly follow a pixel-aligned generation paradigm.
Although monocular depth priors provide useful geometric cues, the Gaussian supports remain constrained by the regular image lattice, limiting their flexibility and adaptation to local surface geometry.
This mismatch often becomes more pronounced under large viewpoint changes, where the pixel-aligned Gaussians may fail to maintain coherent surface-aligned structures, leading to cracks, holes, and unstable geometry in novel-view rendering.
In contrast, InfiniSplat targets surface-aligned Gaussian generation by constructing geometry-guided supports that better follow the local surface layout and decoding Gaussian attributes on these supports through an implicit decoder.
This moves Gaussian placement beyond pixel-aligned supports while preserving geometric regularity, leading to more stable surface organization and more robust large-baseline rendering.

\subsection{Implicit Neural Representations}

Implicit neural representations model signals as functions over query locations and have been widely used for 3D shape, radiance fields, point cloud completion, image representation, and dense prediction~\cite{chen2021learning,saito2019pifu,zhang20233dshape2vecset,yu2026infinidepth,jung2023anyflow,su2023point,xu2026towards}.
LIIF~\cite{chen2021learning} represents an image as a continuous function that can be queried at arbitrary coordinates, enabling super-resolution and flexible output sizes.
InfiniDepth~\cite{yu2026infinidepth} models depth as neural implicit fields, allowing for arbitrary-resolution depth estimation.
InfiniSplat follows the implicit decoding idea but applies it to Gaussian scene generation with explicit geometric organization.
This distinction is important because our output is not a single-channel image or depth signal, but a set of 3D Gaussian primitives whose placement, shape, opacity, and appearance must jointly form a coherent surface-like layout.

Recent feed-forward 3DGS methods also explore query-based Gaussian decoding.
C3G~\cite{an2025c3g} and TokenGS~\cite{ren2026tokengs} learn compact 3D Gaussians from unposed sparse multi-view images by using learnable query tokens to aggregate multi-view features and decode a small set of essential Gaussians.
Such formulations are effective in the multi-view NVS setting, where multiple observations provide cross-view constraints that help learnable queries discover compact and consistent Gaussian layouts.
However, InfiniSplat targets the more under-constrained single-image NVS setting, where these multi-view cues are unavailable.
In this case, learnable query tokens may lack explicit geometric anchors and can struggle to organize Gaussians along coherent scene surfaces under single-view ambiguity.
To address this, InfiniSplat uses geometry-guided support queries instead of freely learned query tokens.
These supports are derived from monocular geometric priors and provide a surface-aware scaffold for Gaussian generation.
On top of these support queries, a query-conditioned implicit decoder predicts Gaussian attributes, preserving the flexibility of query-based decoding while grounding the generation process in explicit geometric structure.
This leads to more constrained, stable, and surface-aligned Gaussian distributions for single-image NVS, achieving higher-quality renderings under different viewpoints.

\section{Method}

\begin{figure*}[ht]
    \centering
    \includegraphics[width=\linewidth]{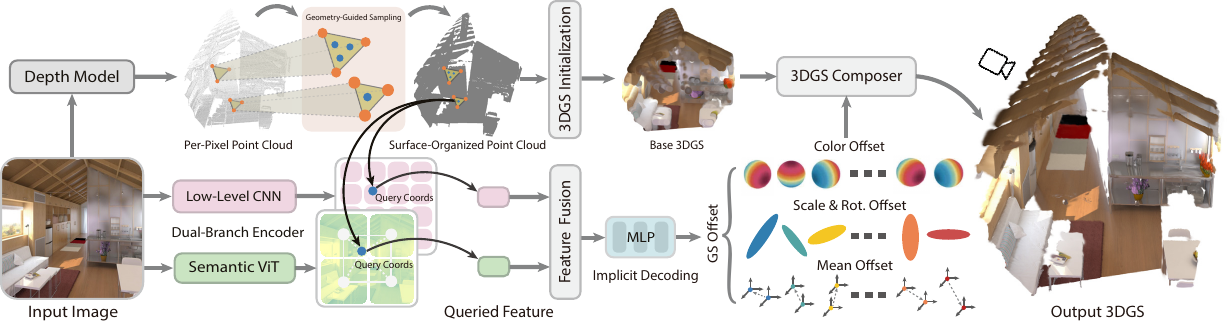}
    \caption{
        \textbf{InfiniSplat pipeline.}
        Given a single input image, InfiniSplat predicts geometry priors and extracts DINO/CNN image features.
        Guided by the predicted depth-induced surface layout, it samples 2D supports and initializes base Gaussians beyond the fixed pixel lattice.
        The implicit Gaussian decoder queries image features at these supports and predicts bounded updates to produce a surface-aligned 3D Gaussian representation for feed-forward novel view synthesis.
        }
        \label{fig:pipeline}
    \vspace{-3mm}
\end{figure*}

\subsection{Method Overview}

InfiniSplat aims to predict a renderable 3D Gaussian scene representation from a single input image in a fully feed-forward manner.
Given an image $\mathbf{I} \in \mathbb{R}^{H \times W \times 3}$, the model outputs a Gaussian set $\mathcal{G} = \{g_i\}_{i=1}^{N}$ without per-scene optimization, where each Gaussian stores position, scale, rotation, color, and opacity.
Our key idea is to shift Gaussian generation from a fixed pixel grid to surface-aligned locations.
This raises two sub-problems: (i)~deciding \emph{where} Gaussians should be placed along scene surfaces, and (ii)~predicting Gaussian parameters at these irregularly distributed locations.
The pipeline accordingly consists of two parallel preprocessing branches feeding into a tightly coupled core of two stages (See Figure~\ref{fig:pipeline}).

\textbf{Parallel preprocessing.}
Two representations are extracted from the input image independently.
A frozen monocular depth model $\Phi_\text{geo}$ predicts a dense depth map $\mathbf{D}$
and camera intrinsics $\hat{\mathbf{K}}$, serving as the geometric scaffold.
A trainable dual-branch encoder $\Phi_\text{img}$ extracts a DINO semantic feature map
$\mathbf{F}_\text{dino}$ and a CNN texture feature map $\mathbf{F}_\text{cnn}$, providing
complementary local conditioning signals for downstream decoding.

\textbf{Coupled core.}
The core comprises two stages whose coupling is central to the method.
(1)~\textit{Geometry-guided sampling}~$\mathcal{Q}$ constructs surface-aligned 2D supports from the predicted depth.
It back-projects the depth map to obtain per-patch 3D surface areas, distributes supports proportionally to area, and initializes each support as a base Gaussian.
This stage determines \emph{where} Gaussians are placed and provides a geometrically meaningful scaffold.
(2)~\textit{Implicit Gaussian decoding}~$\mathcal{D}_\theta$ predicts parameter updates at these supports.
For each support, a shared MLP bilinearly queries $\mathbf{F}_\text{dino}$ and $\mathbf{F}_\text{cnn}$ at the support coordinate, fuses them through a gated mechanism, and predicts bounded offsets to position, scale, rotation, color, and opacity.
Because the decoder is a shared function over query coordinates, it naturally handles the irregular support layout produced by the sampling stage.

The two core stages are coupled by design. Geometry-guided sampling breaks free from the pixel grid but produces an irregular layout that per-pixel regression cannot handle; implicit decoding provides the flexibility to
operate on arbitrary layouts but would lack surface awareness without the geometric scaffold
provided by sampled supports. Each stage necessitates the other: without the decoder, irregular supports cannot be turned
into complete Gaussians; without geometry-guided supports, the decoder reduces to operating on an unstructured point set with no surface prior.

\textbf{Formulation.}
Formally, the pipeline is:
\[
\begin{aligned}
(D,\hat K) &= \Phi_{\mathrm{geo}}(I), \\
(F^{\mathrm{dino}},F^{\mathrm{cnn}}) &= \Phi_{\mathrm{img}}(I), \\
(\mathcal{S},\bar{\mathcal{G}}) &= \mathcal{Q}(D,I,\hat K), \\
\mathcal{G} &= \mathcal{D}_{\theta}(\bar{\mathcal{G}},\mathcal{S},F^{\mathrm{dino}},F^{\mathrm{cnn}}).
\end{aligned}
\]
Here, $\Phi_\text{geo}$ provides the geometric scaffold for support construction
(Section~\ref{sec:representation}).
$\Phi_\text{img}$ extracts complementary semantic and texture features
(Section~\ref{sec:encoder}).
$\mathcal{Q}$ samples a support set $\mathcal{S} = \{q_i\}_{i=1}^{N}$ guided by surface
geometry and initializes the base Gaussian set $\bar{\mathcal{G}}$
(Section~\ref{sec:representation}).
$\mathcal{D}_\theta$ queries $\mathbf{F}_\text{dino}$ and $\mathbf{F}_\text{cnn}$ at each
$q_i$ and predicts bounded updates transforming $\bar{\mathcal{G}}$ into the final Gaussians
$\mathcal{G}$ (Section~\ref{sec:decoding}).

\subsection{Surface-Aligned Gaussian Representation}
\label{sec:representation}
Existing single-image feed-forward 3DGS methods usually strongly bind Gaussian generation locations to the discrete image grid, so their output representations can easily degenerate into splat expansion from pixels or depth points.
In contrast, InfiniSplat constructs a surface-aligned Gaussian representation through geometry-guided sampling, which flexibly initializes base Gaussians under geometric guidance and enables subsequent decoding on supports closer to local surface structures.

\paragraph{Geometry-Guided Sampling.}
Given a dense depth map predicted by a pretrained monocular depth estimation model, e.g., DepthPro~\cite{bochkovskiy2025depth}, we estimate local 3D surface areas from the induced 3D geometry.
Let $p=(x,y)$ denote a pixel location on the image plane, $D(p)$ denote the predicted depth, and $\Pi^{-1}$ denote the back-projection function defined by the estimated intrinsics $\hat K$.
Then the 3D point corresponding to pixel $p$ is
\[
\mathbf{X}(p)=\Pi^{-1}(p,D(p);\hat K).
\]
For a local triangle $t=(p_a,p_b,p_c)$ formed by neighboring pixels, we use its 3D area after back-projection as the sampling weight of this local surface patch:
\[
A_t=\frac{1}{2}
\left\|
\big(\mathbf{X}(p_b)-\mathbf{X}(p_a)\big)
\times
\big(\mathbf{X}(p_c)-\mathbf{X}(p_a)\big)
\right\|_2 .
\]

Before surface sampling, we discard candidate triangles whose vertices exhibit large relative depth variation, preventing faces from spanning foreground--background depth discontinuities.

Based on these triangle areas, we perform area-weighted sampling within the corresponding 2D pixel-grid regions.
Specifically, we normalize the area of each local triangle into a sampling allocation weight:
\[
w_t=\frac{A_t}{\sum_{t'} A_{t'}} .
\]
The local triangles are used to estimate local surface coverage demand, and the resulting $w_t$ determines which local triangles receive more sampled supports.
Local surfaces with larger area or stronger slant are assigned more samples.
We then generate sampled query coordinates $\{q_i\}_{i=1}^{N}$ within the corresponding 2D triangle regions.

\paragraph{Base Gaussian Initialization.}
Each sampled query coordinate is finally initialized as a base Gaussian.
We use the bilinearly sampled values of the input image and depthmap at $q_i$ to initialize its geometry and appearance:
\[
\bar{g}_i=
(\bar{\boldsymbol{\mu}}_i,\bar{\mathbf{s}}_i,\bar{\mathbf{r}}_i,\bar{\mathbf{c}}_i,\bar{\alpha}_i),
\quad
\bar{\boldsymbol{\mu}}_i=\Pi^{-1}(q_i,D(q_i);\hat K),
\quad
\bar{\mathbf{c}}_i=I(q_i).
\]
Here, $\bar{\mathbf{s}}_i$, $\bar{\mathbf{r}}_i$, and $\bar{\alpha}_i$ use a unified base initialization.
At this point, this subsection outputs the geometry-guided sampled support set $\mathcal{S}$ and a base Gaussian set $\bar{\mathcal{G}}=\{\bar{g}_i\}_{i=1}^{N}$, and the subsequent implicit decoder only predicts parameter updates on this base set.

The role of this sampling strategy is not simply to increase the number of points, but to change the support domain of Gaussian generation.
Because the base Gaussians are sampled from depth-induced local surface patches, the decoder can learn Gaussian placement, shape, and appearance on supports that better align with the input geometry.
For large planes, slanted surfaces, object boundaries, and regions with depth variation, this flexible query sampling can reduce the discretization limitation introduced by fixed pixel grid points, making the generated Gaussians easier to organize into spatially coherent structures rather than a set of unrelated point-like splats.

\subsection{Implicit Gaussian Decoding}

InfiniSplat formulates Gaussian decoding as query-conditioned implicit parameter updates to the base Gaussian set.
Here, query-conditioned means that $q_i$ indexes trainable image feature maps to query conditional features, which are combined with the corresponding base Gaussian to predict a parameter update.
Given $\bar{\mathcal{G}}$ and $\mathcal{S}$ from the previous subsection, the decoder predicts Gaussian update for each support.
The whole module contains three sequential steps: feature query, feature fusion, and Gaussian parameter update.

\paragraph{Feature Query.}
\label{sec:encoder}
Feature query extracts local conditioning features for each base Gaussian.
The input image passes through a dual-branch image encoder, where DINO-branch produces a high-level semantic feature map $F^{\mathrm{dino}}$ and CNN-branch generates a low-level appearance feature map $F^{\mathrm{cnn}}$, respectively.
The DINO feature map is obtained by reshaping the final-layer patch tokens and provides part-level semantic consistency; the CNN feature map preserves stronger local texture and edge cues.
For the support coordinate $q_i$ of base Gaussian $\bar{g}_i$, the decoder uses bilinear grid sampling to query the two types of features:

\[
\mathbf{f}^{\mathrm{dino}}_i=\mathcal{B}(F^{\mathrm{dino}},q_i),
\quad
\mathbf{f}^{\mathrm{cnn}}_i=\mathcal{B}(F^{\mathrm{cnn}},q_i),
\]
where $\mathcal{B}$ denotes bilinear sampling.
Similar to the local implicit decoder of InfiniDepth~\citep{yu2026infinidepth}, this step enables the decoder to query image features at sampled query coordinates.
The difference is that we do not use a multi-scale DINO feature pyramid, we use a single-layer DINO feature to provide semantic context and a CNN feature to supplement local appearance details.

\paragraph{Feature Fusion.}
Feature fusion combines the two types of query features into a single Gaussian descriptor.
By default, we first project the CNN feature to the dimension of the DINO feature, and then use channel-wise gated fusion to adaptively balance semantic context and local appearance detail:
\[
\tilde{\mathbf{f}}^{\mathrm{cnn}}_i=P(\mathbf{f}^{\mathrm{cnn}}_i),
\quad
\boldsymbol{\alpha}_i=
\sigma\!\left(W_g[\mathbf{f}^{\mathrm{dino}}_i,\tilde{\mathbf{f}}^{\mathrm{cnn}}_i]\right),
\]
\[
\mathbf{h}_i=
\boldsymbol{\alpha}_i \odot \mathbf{f}^{\mathrm{dino}}_i
+(1-\boldsymbol{\alpha}_i)\odot \tilde{\mathbf{f}}^{\mathrm{cnn}}_i .
\]
Here, $\mathbf{h}_i$ is the fused descriptor finally sent to the Gaussian MLP.
The gate adaptively balances semantic and appearance cues for attribute prediction at each support.

\paragraph{Gaussian Parameter Update.}
\label{sec:decoding}
Although the base Gaussians provide an initial geometry-aware scaffold, their attributes are still coarse and incomplete.
In particular, the base scale, rotation, opacity, and color are not directly optimized for novel-view rendering, and the base mean also needs to be locally adjusted to compensate for errors in the monocular depth prior.
Therefore, we let the implicit decoder predict bounded residual updates of the base Gaussians.
This design preserves the surface-aligned structure introduced by geometry-guided sampling while allowing each Gaussian to adapt its position, shape, and appearance according to local image features.

Specifically, Gaussian parameter update uses an MLP to map $\mathbf{h}_i$ into a Gaussian update $\Delta g_i$, whose channels correspond to 3D placement, scale, rotation, color, and opacity:
\[
\Delta g_i=\mathrm{MLP}_{\theta}(\mathbf{h}_i)
=
(\Delta \mathbf{u}_i,\Delta z_i,\Delta \mathbf{s}_i,\Delta \mathbf{r}_i,\Delta \mathbf{c}_i,\Delta \alpha_i)
\in\mathbb{R}^{14}.
\]
Here, $\Delta \mathbf{u}_i\in\mathbb{R}^{2}$ updates the image-plane support location, $\Delta z_i$ updates the depth direction, and $\Delta \mathbf{s}_i\in\mathbb{R}^{3}$, $\Delta \mathbf{r}_i\in\mathbb{R}^{4}$, $\Delta \mathbf{c}_i\in\mathbb{R}^{3}$, and $\Delta \alpha_i\in\mathbb{R}$ update Gaussian scale, quaternion rotation, color, and opacity, respectively.
The Gaussian composer then applies this update to the corresponding base Gaussian to obtain the final Gaussian primitive.
For position, the composer predicts constrained image-plane displacement and depth displacement near the base support, and then back-projects them to obtain the final 3D mean:
\[
\boldsymbol{\mu}_i
=
\Pi^{-1}\!\left(
\bar{\mathbf{u}}_i+\lambda_{xy}\Delta \mathbf{u}_i,
\bar{d}_i+\lambda_z\Delta z_i;\hat K
\right).
\]
Here, $\bar{\mathbf{u}}_i$ and $\bar{d}_i$ are the base support location and base depth, and $\lambda_{xy}$ and $\lambda_z$ control the spatial update range.
For constrained attributes that must remain in valid ranges, such as scale, color, and opacity, we add updates in the unconstrained space of the corresponding activation following SHARP~\citep{mescheder2025sharp}:
\[
g_i^{(a)}
=
\psi_a\!\left(
\psi_a^{-1}(\bar{g}_i^{(a)})
+\lambda_a \Delta g_i^{(a)}
\right),
\quad
a\in\{\mathrm{scale},\mathrm{color},\mathrm{opacity}\},
\]
where $\psi_a$ is an attribute-specific activation and $\lambda_a$ is the update scale for the corresponding attribute.
The decoder predicts a bounded update around the base value, and the activation ensures that scale, color, and opacity remain valid.
Rotation predicts an orientation update on the base quaternion and is normalized in the subsequent Gaussian transformation.
In this way, the base Gaussian provides stable geometry and color initialization, and the implicit decoder only predicts adaptive placement, shape, and appearance updates according to local image conditions.

The role of this query-conditioned decoding is to turn Gaussian prediction into a shared function over sampled supports and their queried image features.
Because all base Gaussians share the same feature query, fusion module, and Gaussian MLP, the decoder learns a unified update rule, which is helpful for producing more consistent placement, shape, and appearance on neighboring supports.

\subsection{Model Variants and Training Objectives}

\paragraph{Model Variants.}
InfiniSplat contains RGB only and LiDAR conditioned variants under the same framework.
\textbf{InfiniSplat-RGB} is the main setting of this paper; its input is only a single RGB image, and its geometric scaffold comes from DepthPro.
\textbf{InfiniSplat-LiDAR} additionally receives a sparse depth or LiDAR prompt in addition to the RGB image and obtains stronger geometric conditioning through InfiniDepth-Metric~\citep{yu2026infinidepth}.
For each source image, we sample 1,500 sparse depth samples from the source-view depth as prompts.
InfiniDepth-Metric estimates dense depth from the RGB image and these prompts.
The geometry prior remains frozen in both variants, while the dual-branch image branch, feature fusion module, and Gaussian MLP are trainable.
Except for the input form of the geometry prior, the two variants share geometry-guided flexible query sampling, the query-conditioned implicit Gaussian decoder, and the subsequent Gaussian generation pipeline.

\paragraph{Training Objectives.}
We train both models on Hypersim.
Each training sample contains one context view and multiple target views.
The model generates Gaussians from the context view in a feed-forward manner and renders images under the target cameras for supervision:
\[
\hat{I}_t=\mathcal{R}(\mathcal{G};K_t,T_t),
\]
where $\mathcal{R}$ denotes the Gaussian renderer, and $K_t$ and $T_t$ denote the intrinsics and camera pose of the target view, respectively.
The training objective consists of rendering supervision and Gaussian regularization.
Rendering supervision includes an RGB reconstruction loss and a perceptual loss following SHARP~\citep{mescheder2025sharp}:
\[
\mathcal{L}_{\mathrm{rgb}}
=
\frac{1}{|\mathcal{T}|}
\sum_{t\in\mathcal{T}}
\left\|\hat{I}_t-I_t\right\|_1 ,
\]
\[
\mathcal{L}_{\mathrm{perc}}
=
\frac{1}{|\mathcal{T}|}
\sum_{t\in\mathcal{T}}
\sum_l
\left(
\left\|\phi_l(\hat{I}_t)-\phi_l(I_t)\right\|_2^2
+\gamma
\left\|G_l(\hat{I}_t)-G_l(I_t)\right\|_F^2
\right),
\]
where $\mathcal{T}$ denotes the target views.
This perceptual loss combines a feature-space distance and a Gram-matrix distance; $\phi_l$ is the frozen VGG feature at layer $l$, and $G_l$ denotes the corresponding feature Gram matrix.
The RGB loss provides pixel-level reconstruction constraints, while the perceptual loss constrains high-level feature similarity and texture statistics so that rendered results preserve better perceptual quality on large-baseline target views.

In addition to the rendering losses, we use Gaussian regularization to stabilize Gaussian attributes.
This term mainly constrains the valid range of Gaussian scales and encourages scale and opacity to remain locally consistent across neighboring sampled supports:
\[
\mathcal{L}_{\mathrm{reg}}
=
\lambda_{\mathrm{scale}}\mathcal{L}_{\mathrm{scale}}
+\lambda_{\mathrm{smooth}}\mathcal{L}_{\mathrm{smooth}} .
\]
Scale regularization constrains Gaussian scales within an effective range, avoiding overly large splats or degeneration into extremely small Gaussians.
Let $\mathbf{s}_i$ denote the scale of the $i$-th Gaussian:
\[
\mathcal{L}_{\mathrm{scale}}
=
\frac{1}{N}\sum_i
\left[
\mathrm{ReLU}(\ell_{\min}-\log \mathbf{s}_i)
+\mathrm{ReLU}(\log \mathbf{s}_i-\ell_{\max})
\right].
\]
Local smoothness regularization constrains scale and opacity to vary smoothly across neighboring sampled supports.
Let $\mathcal{N}(i)$ denote the neighboring supports of $q_i$:
\[
\mathcal{L}_{\mathrm{smooth}}
=
\frac{1}{N}\sum_i
\frac{1}{|\mathcal{N}(i)|}
\sum_{j\in\mathcal{N}(i)}
\left(
\left\|\log \mathbf{s}_i-\log \mathbf{s}_j\right\|_1
+|\alpha_i-\alpha_j|
\right).
\]
These Gaussian regularization terms constrain the valid range and local consistency of Gaussian attributes, allowing rendering supervision to more stably optimize the same Gaussian set.

The overall training objective is
\[
\mathcal{L}
=
\lambda_{\mathrm{rgb}}\mathcal{L}_{\mathrm{rgb}}
+\lambda_{\mathrm{perc}}\mathcal{L}_{\mathrm{perc}}
+\lambda_{\mathrm{reg}}\mathcal{L}_{\mathrm{reg}}.
\]
The perceptual loss is only applied to target views, while the context view is mainly used for constraints that stabilize the generated representation.
Thus, the main image supervision comes directly from novel-view rendering, while Gaussian regularization only serves as a training constraint for stabilizing the explicit Gaussian representation.

\section{Experiments}

\subsection{Experimental Setup}

\begin{table*}[t]
\centering
\small
\setlength{\tabcolsep}{2.4pt}
\caption{Quantitative evaluation on zero-shot novel-view synthesis. PSNR, SSIM, and LPIPS are reported as separate metric columns. Higher is better for PSNR and SSIM, while lower is better for LPIPS. The best and second-best results are highlighted in \protect\colorbox{green!30!yellow!30}{green} and \protect\colorbox{yellow!30}{yellow}, respectively, within each setting block.}
\label{tab:main_comparison}
\resizebox{\textwidth}{!}{%
\begin{tabular}{lccccccccccccccc}
\toprule
Method
& \multicolumn{3}{c}{ETH3D}
& \multicolumn{3}{c}{ScanNet++}
& \multicolumn{3}{c}{Tanks-and-Temples}
& \multicolumn{3}{c}{DL3DV}
& \multicolumn{3}{c}{Avg.} \\
\cmidrule(lr){2-4}
\cmidrule(lr){5-7}
\cmidrule(lr){8-10}
\cmidrule(lr){11-13}
\cmidrule(lr){14-16}
& PSNR $\uparrow$ & SSIM $\uparrow$ & LPIPS $\downarrow$
& PSNR $\uparrow$ & SSIM $\uparrow$ & LPIPS $\downarrow$
& PSNR $\uparrow$ & SSIM $\uparrow$ & LPIPS $\downarrow$
& PSNR $\uparrow$ & SSIM $\uparrow$ & LPIPS $\downarrow$
& PSNR $\uparrow$ & SSIM $\uparrow$ & LPIPS $\downarrow$ \\
\midrule
\multicolumn{16}{l}{\textit{RGB-only / monocular methods}} \\
LVSM & 17.306 & 0.532 & 0.488 & 13.177 & 0.499 & 0.538 & 13.696 & 0.394 & 0.531 & 14.249 & 0.311 & 0.518 & 14.607 & 0.434 & 0.519 \\
LagerNVS & 18.130 & 0.598 & 0.462 & 16.749 & 0.634 & 0.395 & \secondcell{16.844} & 0.661 & 0.371 & 18.118 & 0.510 & 0.571 & 17.460 & 0.601 & 0.450 \\
Flash3D & 18.546 & \secondcell{0.874} & \secondcell{0.239} & 17.375 & 0.782 & 0.350 & 15.952 & \secondcell{0.674} & 0.326 & 17.502 & 0.622 & 0.395 & 17.344 & 0.738 & 0.328 \\
Flash3D-DepthPro & 17.917 & 0.853 & 0.257 & 17.561 & 0.775 & 0.367 & 12.168 & 0.652 & 0.410 & 17.602 & 0.642 & 0.405 & 16.312 & 0.731 & 0.360 \\
SHARP & \secondcell{19.046} & 0.868 & 0.241 & \secondcell{20.801} & \secondcell{0.838} & \secondcell{0.273} & 15.750 & \secondcell{0.674} & \secondcell{0.310} & \secondcell{18.305} & \secondcell{0.652} & \secondcell{0.370} & \secondcell{18.475} & \secondcell{0.758} & \secondcell{0.299} \\
\textbf{InfiniSplat-RGB} & \bestcell{20.531} & \bestcell{0.895} & \bestcell{0.220} & \bestcell{22.240} & \bestcell{0.864} & \bestcell{0.270} & \bestcell{17.118} & \bestcell{0.695} & \bestcell{0.306} & \bestcell{21.685} & \bestcell{0.772} & \bestcell{0.310} & \bestcell{20.394} & \bestcell{0.806} & \bestcell{0.277} \\
\midrule
\multicolumn{16}{l}{\textit{RGB+DepthSensor / LiDAR-conditioned methods}} \\
ADGaussian & \secondcell{12.752} & \secondcell{0.749} & \secondcell{0.337} & \secondcell{10.474} & \secondcell{0.565} & \secondcell{0.506} & \secondcell{13.523} & \secondcell{0.604} & \secondcell{0.412} & -- & -- & -- & \secondcell{12.249} & \secondcell{0.639} & \secondcell{0.418} \\
\textbf{InfiniSplat-LiDAR} & \bestcell{25.880} & \bestcell{0.946} & \bestcell{0.151} & \bestcell{24.148} & \bestcell{0.896} & \bestcell{0.237} & \bestcell{17.617} & \bestcell{0.712} & \bestcell{0.287} & -- & -- & -- & \bestcell{22.548} & \bestcell{0.851} & \bestcell{0.225} \\
\bottomrule
\end{tabular}
}
\end{table*}

\begin{figure*}[ht]
    \centering
    \includegraphics[width=0.91\linewidth]{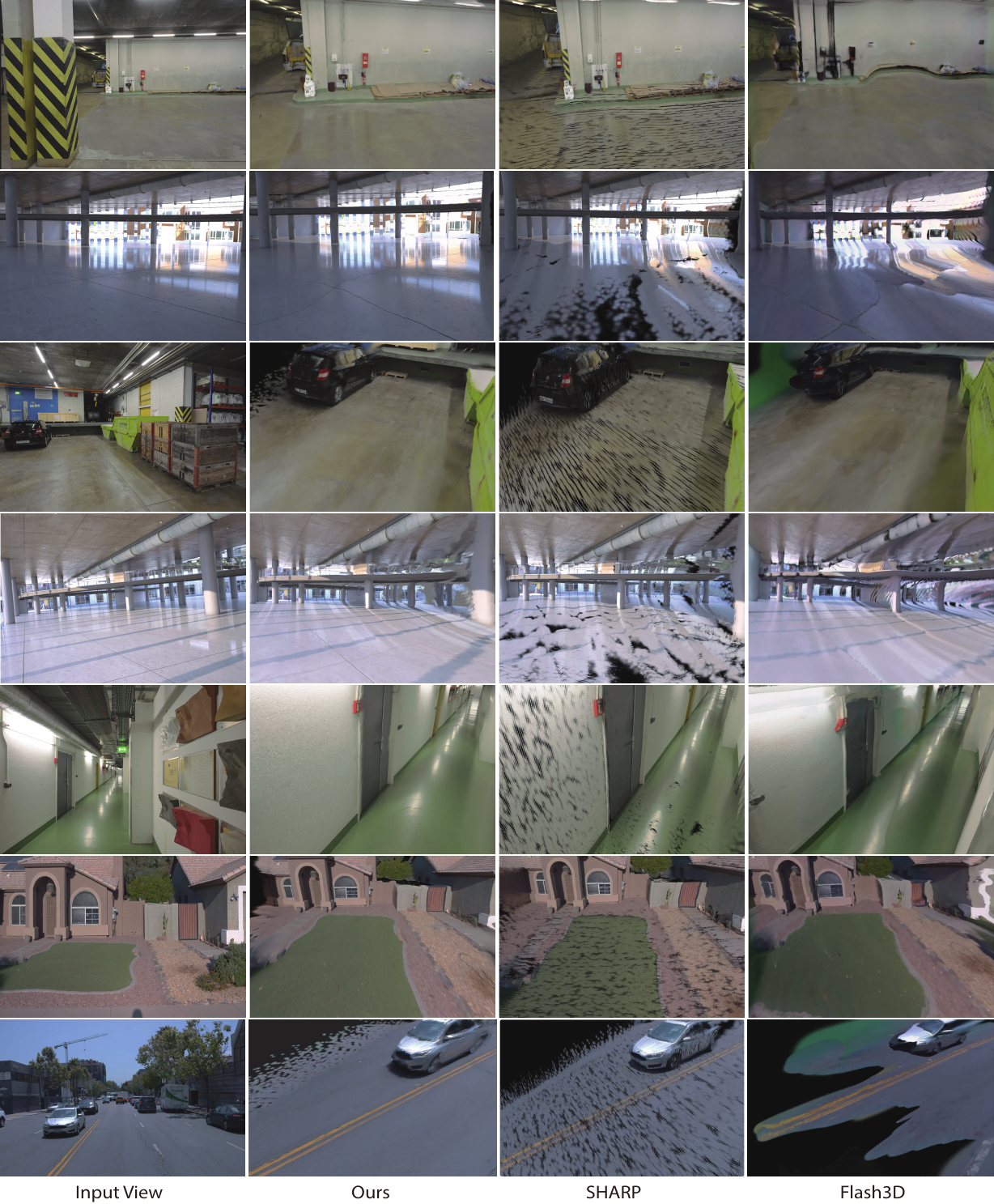}
    \caption{\textbf{RGB qualitative comparison under large viewpoint changes}. InfiniSplat-RGB produces fewer visible cracks and more coherent scene structures than feed-forward baselines, especially on large planar regions and object boundaries.}
\label{fig:qual_rgb}
\end{figure*}

\begin{figure*}[t]
    \centering
    \includegraphics[width=0.90\linewidth]{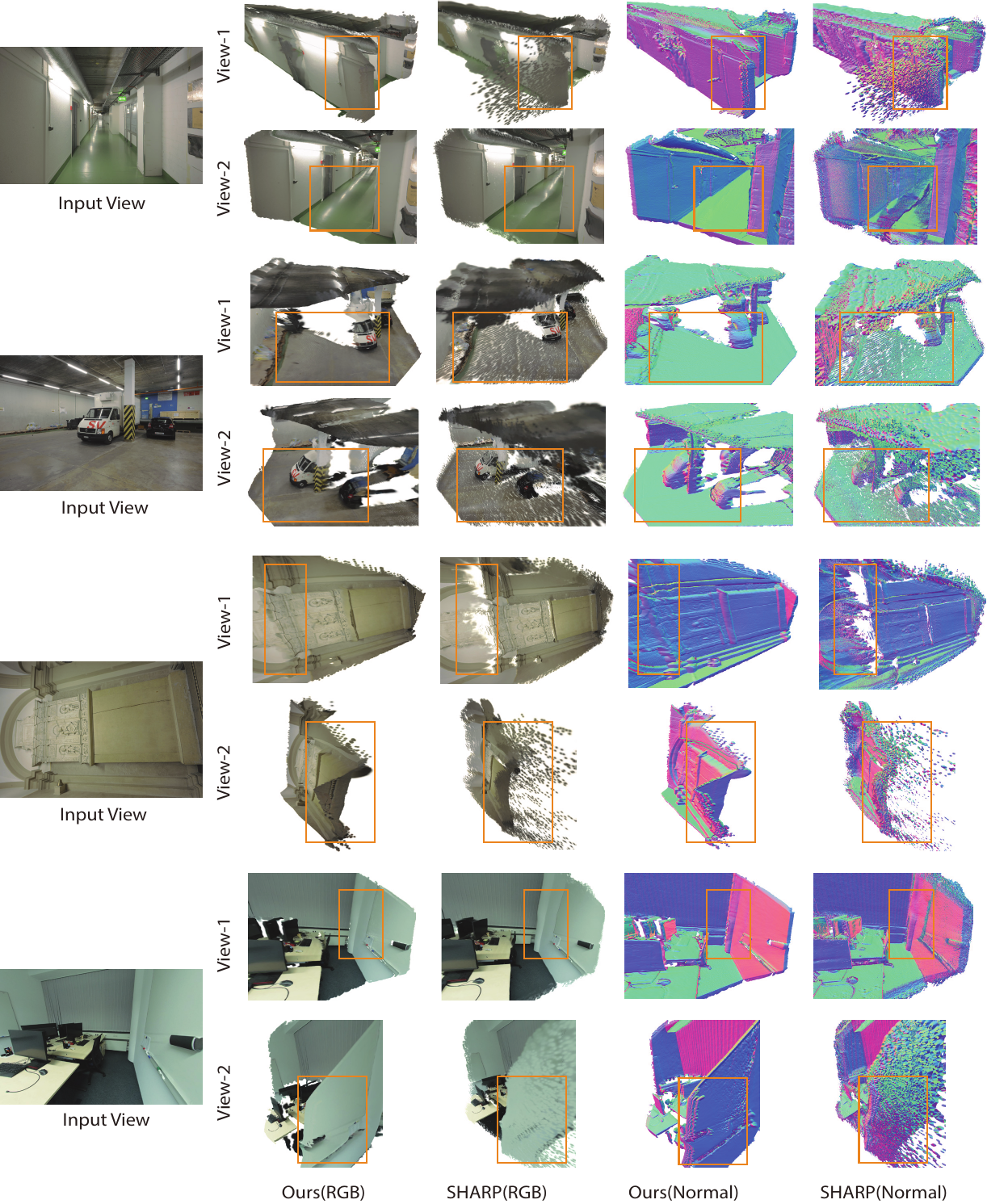}
    \caption{\textbf{Surface and normal comparison}. InfiniSplat produces cleaner normal maps and more coherent surface organization than SHARP, with fewer cracks and fragmented regions under side-view rendering.}
    \label{fig:qual_normal}
\end{figure*}

\begin{figure*}[t]
    \centering
    \includegraphics[width=0.945\linewidth]{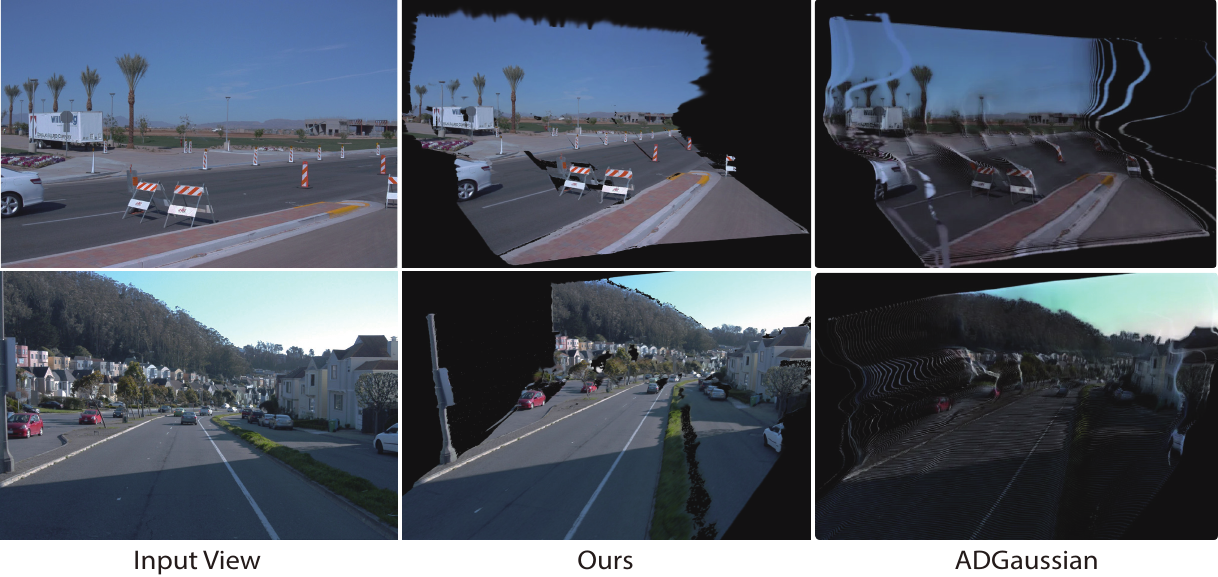}
    \vspace{-3mm}
    \caption{
    \textbf{Qualitative comparison between InfiniSplat-LiDAR and ADGaussian.}
    With sparse LiDAR points as additional input, InfiniSplat-LiDAR produces substantially better novel-view rendering quality than ADGaussian. More examples are included in the supp. video.
    }
    \label{fig:qual_lidar}
\end{figure*}

\begin{figure*}[t]
    \centering
    \includegraphics[width=0.945\linewidth]{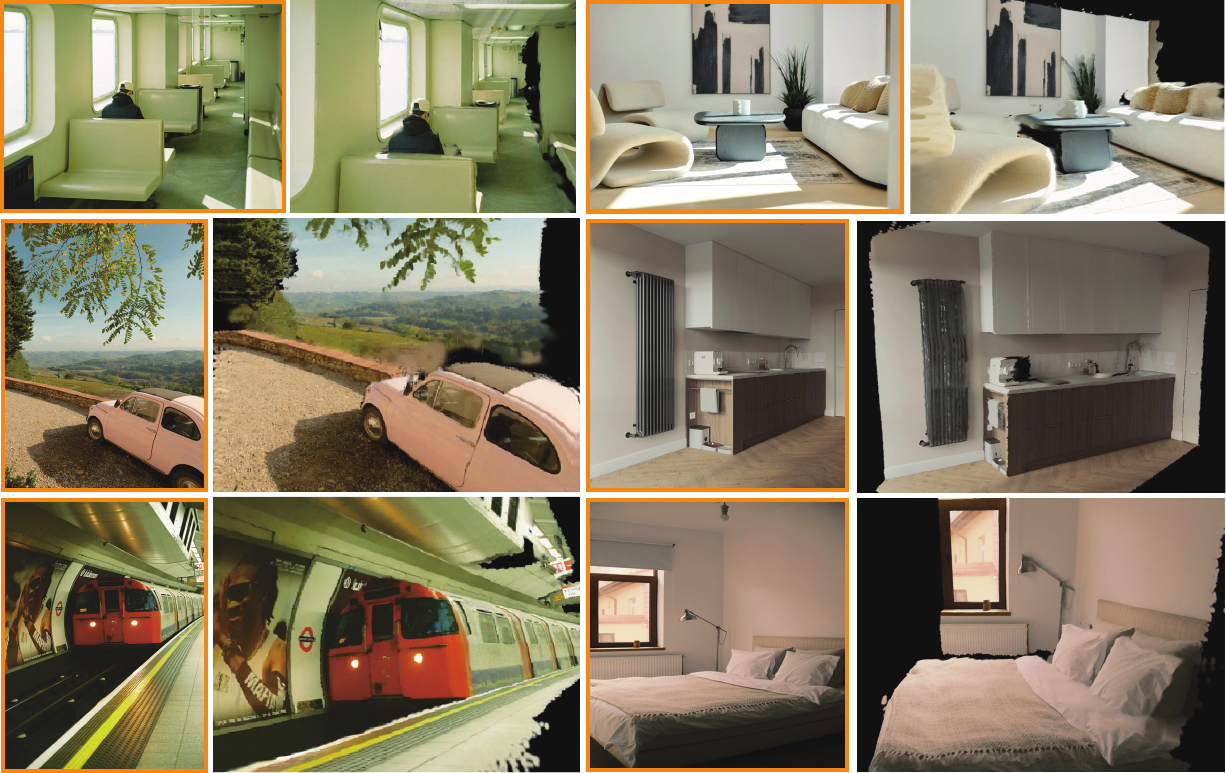}
    \vspace{-2.5mm}
    \caption{\textbf{Qualitative results of InfiniSplat on in-the-wild images}. The model demonstrates strong generalization capabilities, producing detailed and coherent 3D reconstructions even in challenging scenarios with complex geometry and varying lighting conditions.}
    \label{fig:wild}
\end{figure*}

\paragraph{Datasets.}
We evaluate InfiniSplat under a cross-dataset zero-shot novel-view synthesis setting.
Given a single source image, each method directly synthesizes a target view and is evaluated against the ground-truth target image on the same source-target pairs.
We use four real-scene datasets with metric poses: ETH3D~\citep{schops2017multi}, ScanNet++~\citep{yeshwanth2023scannet++}, Tanks and Temples~\citep{knapitsch2017tanks}, and DL3DV~\citep{ling2024dl3dv}.
None of the evaluation datasets is used for training.

For each multi-view dataset, we generate candidate source-target pairs from trajectory windows within each scene or sequence.
Each trajectory window contains 10 consecutive views, from which we enumerate directed source-target candidates.
To ensure sufficient common visibility between the target view and the input view, we require the source-target camera frustum overlap to be greater than $60\%$ and filter out excessive rotations or backward-facing views.
Each dataset contains 512 source-target pairs, covering a range of camera baselines from nearby views to more challenging large-baseline viewpoint changes.
The 512 pairs are selected using a fixed random seed, targeting 128 pairs in each of four camera-baseline ranges: $[0,0.5)$, $[0.5,1)$, $[1,2)$, and $[2,\infty)$ meters.
A scene-level sampling cap is used during the initial selection to reduce overrepresentation by individual scenes.
For evaluation, we preserve the original aspect ratio of each dataset and align the synthesized target views to dataset-specific target sizes before computing metrics: ETH3D uses $2016\times1344$, ScanNet++ uses $1536\times1024$, and Tanks-and-Temples and DL3DV use $1920\times1080$.
Quantitative evaluation and qualitative rendering use the same target sizes.

For datasets with reference depth, we project the source-view geometry into the target view to obtain an initial visibility map, which is further refined using simple image-space morphological operations into a complete frustum-visible mask.
The same evaluation mask is applied to all compared methods.

\paragraph{Baselines and Metrics.}
We evaluate both the RGB-only setting and the RGB+DepthSensor setting.
In the RGB-only setting, we compare with SHARP~\citep{mescheder2025sharp}, the official Flash3D~\citep{szymanowicz2025flash3d}, Flash3D-DepthPro, LVSM~\citep{jin2024lvsm}, and LagerNVS~\citep{szymanowicz2026lagernvs}.
InfiniSplat-RGB and SHARP use DepthPro, while Flash3D-DepthPro replaces the depth model of the official Flash3D with DepthPro to control for the depth backend.
LVSM and LagerNVS provide single-view image-to-image NVS baselines.
In the RGB+DepthSensor setting, we compare with ADGaussian~\citep{song2025adgaussian}.
ADGaussian is a single-image feed-forward 3DGS baseline with source depth input.
We report PSNR, SSIM, and LPIPS~\citep{zhang2018unreasonable}; PSNR and SSIM measure reconstruction fidelity, while LPIPS measures perceptual similarity.

\paragraph{Implementation Details.}
The two InfiniSplat variants are trained on Hypersim~\citep{roberts2021hypersim}.
Each training sample contains one context view and three target views, and each input image samples $1.5\mathrm{M}$ Gaussian supports.
The image branch uses a DINOv3 ViT-L/16 backbone~\citep{simeoni2025dinov3} and a 128-channel CNN.
This CNN branch begins with a $7\times7$ convolution with stride 2, followed by eight residual blocks with $3\times3$ kernels.
The resulting multi-scale features are fused by a $3\times3$ convolution and projected by a final $1\times1$ convolution to a feature map with $C=128$ channels.
The loss weights are set to $\lambda_{\mathrm{rgb}}=1$, $\lambda_{\mathrm{perc}}=1$, and $\lambda_{\mathrm{reg}}=0.1$.
The Gram-matrix term in the perceptual loss uses $\gamma=10$ and is applied only to target views.
For Gaussian regularization, we use $\lambda_{\mathrm{scale}}=0.1$ and $\lambda_{\mathrm{smooth}}=0.1$, and set the log-scale bounds to $[\ell_{\min},\ell_{\max}]=[-8,-3]$ in normalized Gaussian scale space.
Both variants are trained on 8 NVIDIA H20 GPUs for about $100\mathrm{K}$ steps using AdamW, with a learning rate of $5\times10^{-5}$ and a per-GPU batch size of 1.
Unless otherwise specified, all ablations use the same training setting and evaluation protocol.

\subsection{Quantitative Comparison}

Table~\ref{tab:main_comparison} reports the quantitative comparison between InfiniSplat and the baselines.
The table reports PSNR, SSIM, and LPIPS as separate metric columns.
InfiniSplat-RGB achieves the highest PSNR, highest SSIM, and lowest LPIPS on all four RGB-only datasets.
On the four-dataset average, InfiniSplat-RGB reaches $20.394/0.806/0.277$, improving over SHARP by $+1.919$ PSNR and $+0.048$ SSIM while reducing LPIPS by $0.022$.
Compared with Flash3D, Flash3D-DepthPro, and LagerNVS, InfiniSplat-RGB improves the average PSNR by $+3.050$, $+4.082$, and $+2.934$, respectively, while reducing average LPIPS by $0.051$, $0.083$, and $0.173$.
These results show that the improvement of InfiniSplat does not come from a single dataset or a single baseline, but remains consistent across indoor scans, large outdoor scenes, and DL3DV open-world scenes.

Per-dataset results further reflect the structural stability under the large-baseline setting.
Compared with SHARP, InfiniSplat-RGB improves PSNR by $+1.485$, $+1.439$, $+1.368$, and $+3.380$ on ETH3D, ScanNet++, Tanks-and-Temples, and DL3DV, respectively.
Compared with Flash3D, the corresponding PSNR gains are $+1.985$, $+4.865$, $+1.166$, and $+4.183$.
The LPIPS improvement on Tanks-and-Temples is small, but InfiniSplat-RGB still maintains higher PSNR and SSIM.

The LiDAR-conditioned setting further shows that the same implicit Gaussian decoding framework remains effective when the input provides more reliable geometric conditions.
On ETH3D, ScanNet++, and Tanks-and-Temples with source depth, InfiniSplat-LiDAR achieves an average score of $22.548/0.851/0.225$, improving over ADGaussian by $+10.299$ PSNR and $+0.212$ SSIM while reducing LPIPS by $0.193$.
Since DL3DV does not provide usable source-depth input, we do not evaluate it in the LiDAR-conditioned setting.
Overall, the RGB-only and LiDAR-conditioned results together indicate that the improvement comes from the Gaussian representation and decoding mechanism itself, rather than from a particular depth source.

These comparison results support the core claim of this paper: single-image feed-forward 3DGS can move beyond pixel-aligned splat expansion toward a more coherent scene representation.
Compared with SHARP and Flash3D, InfiniSplat moves Gaussian generation from fixed pixel centers to more flexible sampled supports through geometry-guided sampled supports and query-conditioned implicit Gaussian decoding.
Therefore, across multiple cross-dataset real scenes, InfiniSplat can generate more stable target-view renderings and achieves a better balance between structure and perceptual quality.

\subsection{Qualitative Comparison}

Quantitative metrics only partially reflect the quality differences of single-image 3DGS; the main focus of this paper is whether the generated Gaussians remain organized into reasonable surfaces when the viewpoint clearly deviates from the source view.
Figure~\ref{fig:qual_rgb} shows an RGB rendering comparison.
We select examples with large viewpoint shifts and compare the novel-view renderings of InfiniSplat-RGB, SHARP, and Flash3D from the same source view.
SHARP usually preserves strong texture sharpness on nearby views, but under large side-view changes it tends to expose tearing, disconnection, and local drift on large planar regions and object boundaries.
Flash3D is more likely to show global blur, geometric stretching, and structural drift.
In contrast, InfiniSplat-RGB usually preserves more coherent large planes, more stable object boundaries, and fewer holes and cracks.

To observe the Gaussian layout more directly, Figure~\ref{fig:qual_normal} further shows RGB rendering and rendered normal visualization.
RGB rendering can sometimes hide geometric artifacts behind texture sharpness, while normal visualization directly exposes whether the Gaussians form coherent surfaces in space.
The normal visualizations of SHARP often show fragmented surfaces, broken boundaries, and locally fragmented splats.
In contrast, InfiniSplat produces smoother and more coherent rendered normals, especially on walls, floors, columns, vehicles, and building facades.
This visualization complements the quantitative results in Table~\ref{tab:main_comparison}, showing that our advantage is not only an improvement in image metrics, but also comes from a more stable surface-aligned Gaussian representation.

We also compare InfiniSplat-LiDAR with ADGaussian in the RGB+DepthSensor setting.
As shown in Figure~\ref{fig:qual_lidar}, InfiniSplat-LiDAR produces substantially better novel-view rendering quality than ADGaussian, with fewer cracks and holes and more coherent surfaces under larger viewpoint changes.
This further supports the effectiveness of our proposed method.

Additionally, we show qualitative results of InfiniSplat on in-the-wild images in Figure~\ref{fig:wild}. This demonstrates the strong generalization capabilities of our model, producing detailed and coherent 3D reconstructions even in challenging scenarios with complex geometry and varying lighting conditions.
\subsection{Ablation Study}

\begin{table*}[t]
\centering
\setlength{\tabcolsep}{5pt}
\caption{Ablation study on representative zero-shot datasets. PSNR, SSIM, and LPIPS are reported as separate metric columns. Higher is better for PSNR and SSIM, while lower is better for LPIPS. Best results are highlighted in \protect\colorbox{green!30!yellow!30}{green}.}
\label{tab:ablation}
\vspace{-1mm}
\begin{tabular}{lcccccc}
\toprule
Method
& \multicolumn{3}{c}{ETH3D}
& \multicolumn{3}{c}{ScanNet++} \\
\cmidrule(lr){2-4}
\cmidrule(lr){5-7}
& PSNR $\uparrow$ & SSIM $\uparrow$ & LPIPS $\downarrow$
& PSNR $\uparrow$ & SSIM $\uparrow$ & LPIPS $\downarrow$ \\
\midrule
Full model & \bestcell{20.531} & \bestcell{0.895} & \bestcell{0.220} & \bestcell{22.240} & \bestcell{0.864} & \bestcell{0.270} \\
w/o learned updates (base GS only) & 18.819 & 0.877 & 0.237 & 19.861 & 0.833 & 0.298 \\
w/o DINO & 11.465 & 0.768 & 0.336 & 12.501 & 0.655 & 0.431 \\
w/o CNN & 18.646 & 0.873 & 0.240 & 20.368 & 0.833 & 0.293 \\
w/o Gaussian regularization & 19.741 & 0.884 & 0.231 & 20.717 & 0.836 & 0.299 \\
w/o Geometry-guided Sampling & 19.934 & 0.888 & 0.227 & 21.576 & 0.855 & 0.274 \\
w/o Implicit Decoder & 18.678 & 0.873 & 0.240 & 20.811 & 0.833 & 0.290 \\
\bottomrule
\end{tabular}
\end{table*}

\begin{figure}[t]
    \centering
    \includegraphics[width=\linewidth]{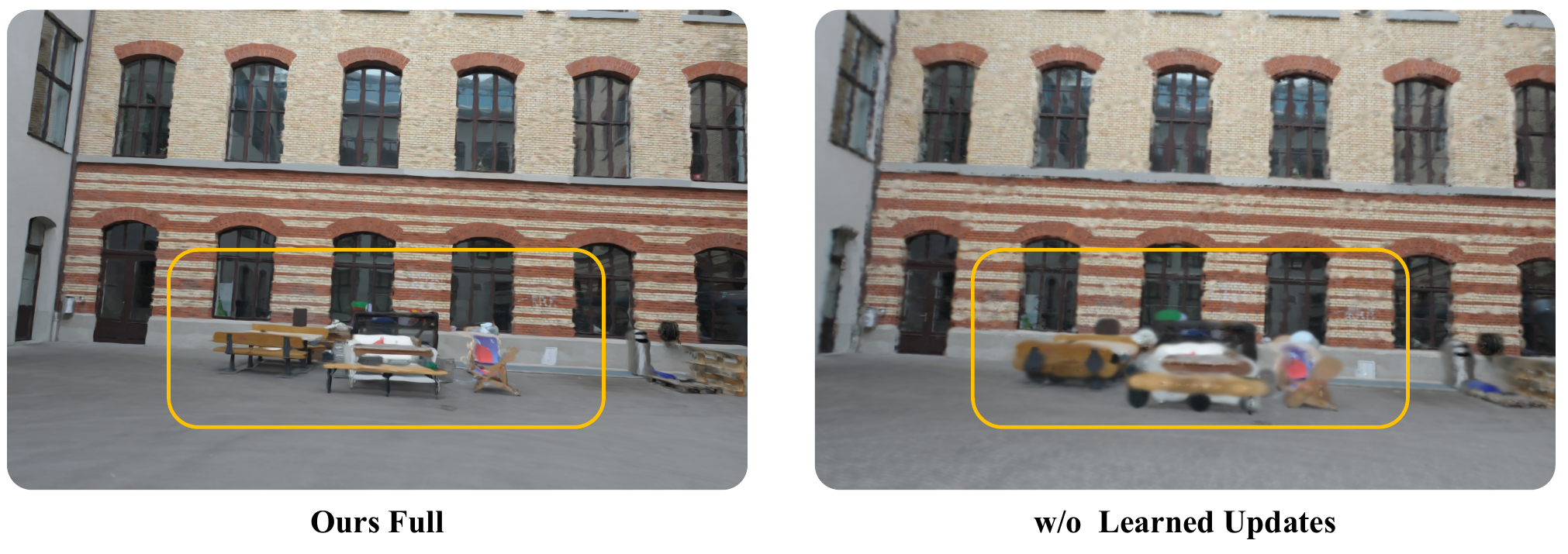}
    \vspace{-4mm}
    \caption{Qualitative effect of learned Gaussian updates. Without learned updates, the model only renders base Gaussians initialized from geometry and sampled color. The result preserves coarse scene structure but remains blurry and lacks local texture and boundary details.}
    \label{fig:qual_ablation_updates}
\end{figure}

\begin{figure}[t]
    \centering
    \includegraphics[width=\linewidth]{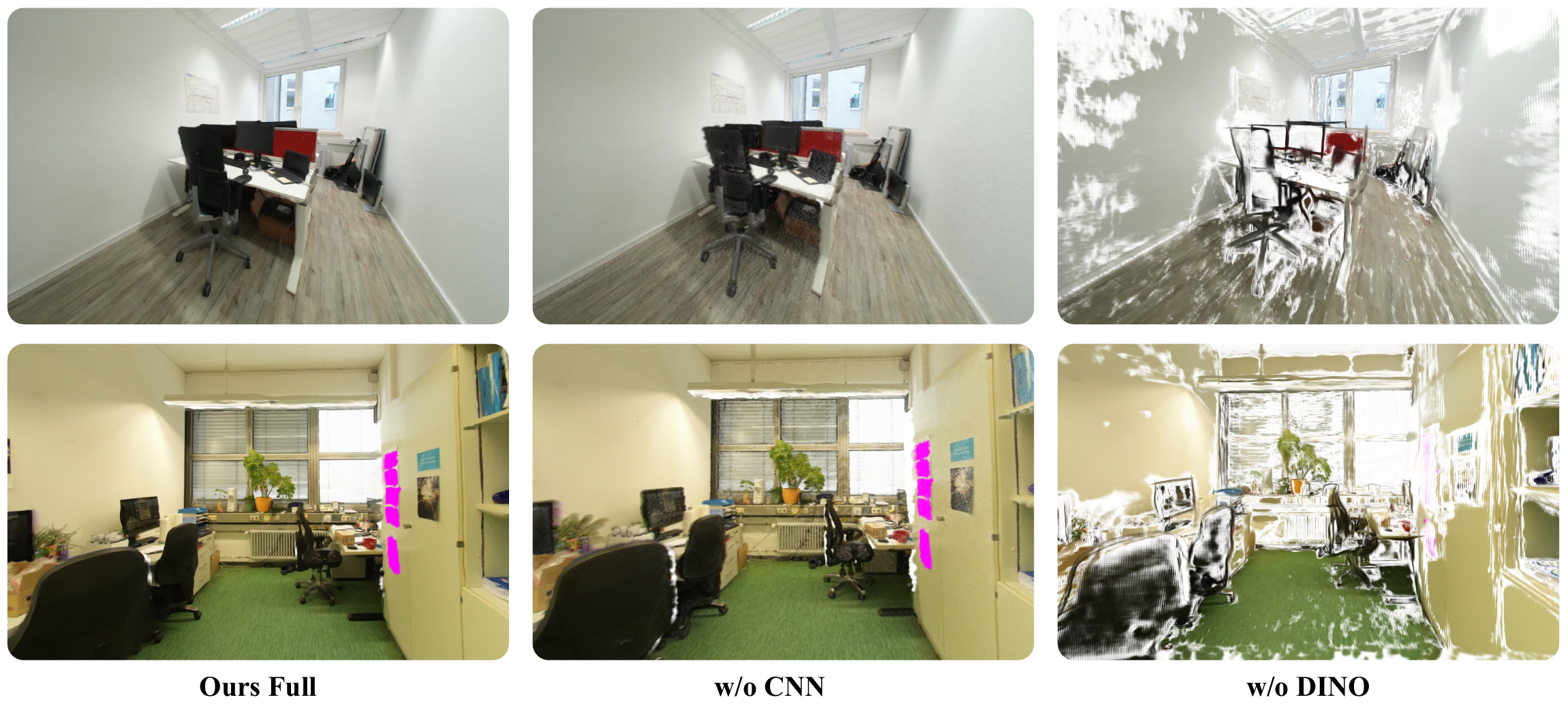}
    \vspace{-4mm}
    \caption{Qualitative effect of image feature branches. Removing the CNN branch weakens low-level appearance and boundary details, producing blurrier renderings. Removing the DINO branch is more destructive: the model loses its main semantic backbone and fails to generalize, producing large holes and missing structures.}
    \label{fig:qual_ablation_image_branches}
\end{figure}

\begin{figure}[t]
    \centering
    \includegraphics[width=\linewidth]{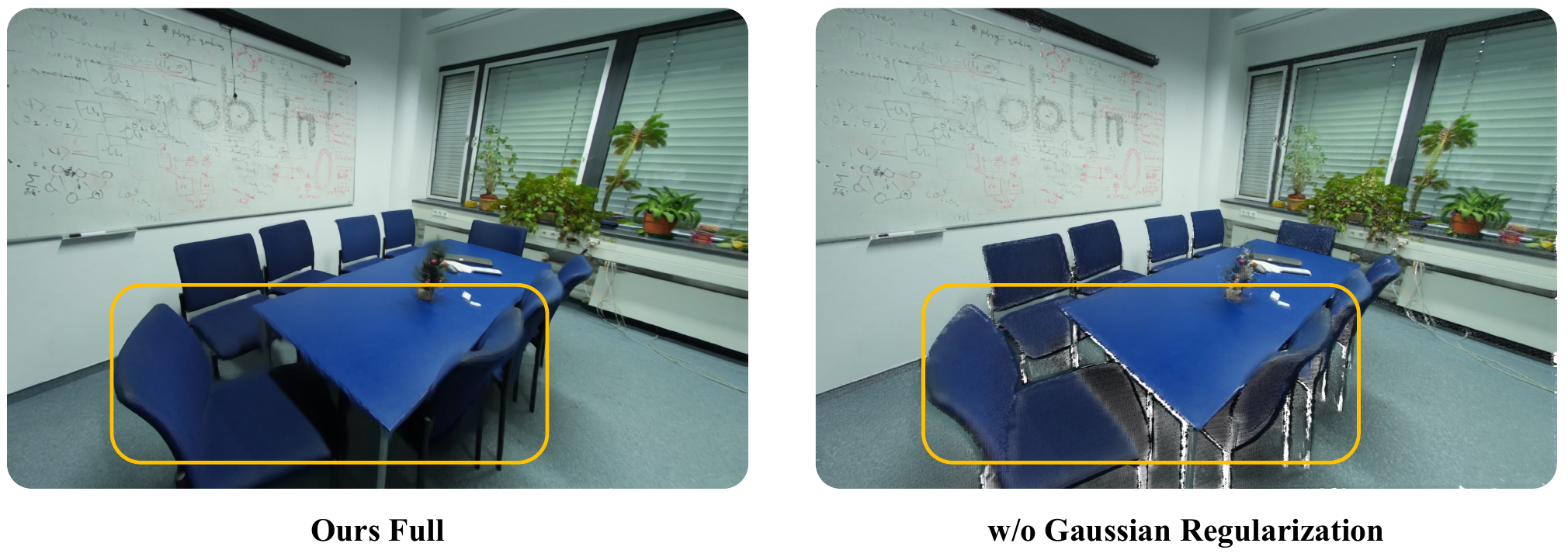}
    \vspace{-4mm}
    \caption{Qualitative effect of Gaussian regularization. Removing Gaussian regularization produces degenerate Gaussian attributes, leading to locally thin, transparent, and unstable regions in the rendered target view.}
    \label{fig:qual_ablation_regularization}
\end{figure}

\begin{figure}[t]
    \centering
    \includegraphics[width=\linewidth]{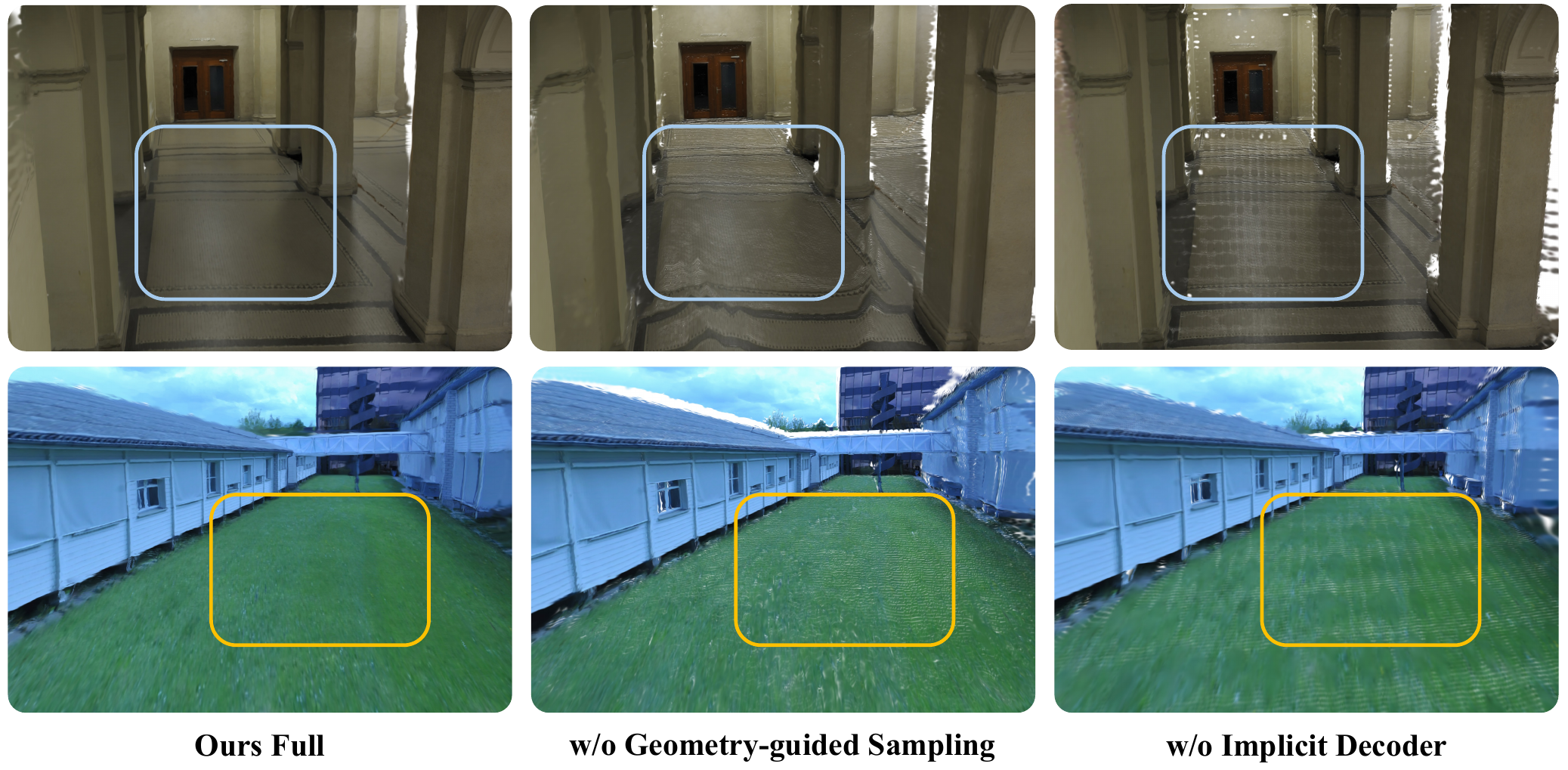}
    \vspace{-4mm}
    \caption{Qualitative effect of geometry-guided sampling and implicit decoding. Replacing geometry-guided sampled supports with pixel-aligned supports introduces cracks under large viewpoint changes. Replacing the implicit decoder with a DPT decoder further amplifies these discontinuities, producing more severe holes and surface breaks.}
    \label{fig:qual_ablation_sampling_decoder}
\end{figure}

Table~\ref{tab:ablation} analyzes the key designs of InfiniSplat-RGB on ETH3D and ScanNet++.
The table reports PSNR, SSIM, and LPIPS as separate metric columns.
In addition to these component ablations, we further study the effect of sampled support budget in Table~\ref{tab:support_budget} to analyze the trade-off between rendering quality and inference cost.

We consider the following component ablation settings.
\textit{w/o learned updates} keeps only the base Gaussians initialized from depth and color, without predicting learned Gaussian updates.
\textit{w/o DINO} removes the DINO branch and keeps only CNN low-level image features.
\textit{w/o CNN} removes the low-level CNN branch and keeps only the DINO feature condition.
\textit{w/o Gaussian regularization} uses only RGB reconstruction and perceptual losses for training, removing Gaussian regularization.
\textit{w/o Geometry-guided Sampling} replaces geometry-guided sampled supports with pixel-aligned supports while keeping the subsequent implicit decoding.
\textit{w/o Implicit Decoder} removes the query-conditioned implicit decoder and instead uses a pixel-aligned Dense Prediction Transformer (DPT) decoder~\citep{ranftl2021vision} to predict Gaussian parameters.

\paragraph{Learned Updates.}
Removing learned Gaussian updates causes significant degradation.
The base-GS-only variant keeps only geometry-guided initialization and sampled color; its PSNR drops by $1.712$ on ETH3D and by $2.379$ on ScanNet++.
LPIPS also degrades from $0.220$ to $0.237$ on ETH3D and from $0.270$ to $0.298$ on ScanNet++.
This shows that the model does not simply render a lifted depth map, but relies on learned Gaussian placement, shape, and appearance updates to form a usable 3DGS representation.
The qualitative results in Figure~\ref{fig:qual_ablation_updates} further support this conclusion.
Without learned updates, the model only renders base Gaussians initialized from geometry and sampled color, so the rendering remains close to initialization.
Although the main scene structure is still visible, textures and local boundaries are noticeably blurry, indicating the lack of detail compensation from learned Gaussian attributes.

\paragraph{Dual-Branch Image Encoder.}
DINO and CNN provide complementary constraints.
Removing the DINO branch in the image encoder causes clear degradation on both datasets, especially on ScanNet++, where PSNR decreases from $22.240$ to $12.501$ and LPIPS increases from $0.270$ to $0.431$.
This suggests that high-level semantic features are especially important for transferring from Hypersim to more complex real indoor scenes.
Figure~\ref{fig:qual_ablation_image_branches} shows a consistent qualitative trend: without DINO, the model is prone to large holes and missing structures in cross-dataset scenes, indicating a clear failure of scene-level generalization.
Removing the CNN branch also leads to a stable drop; on ScanNet++, PSNR decreases to $20.368$ and LPIPS increases to $0.293$.
This indicates that low-level image features remain necessary for local appearance and geometric details.
Qualitatively, \textit{w/o CNN} produces blurrier renderings and weaker fine textures, boundaries, and local geometric details.
Together, these results show that the DINO branch mainly provides robust high-level scene conditions, while the CNN branch preserves fine-scale local information required for accurate Gaussian attribute prediction.

\paragraph{Gaussian Regularization.}
Removing Gaussian regularization decreases results on both datasets; for example, on ScanNet++ PSNR drops from $22.240$ to $20.717$ and LPIPS increases from $0.270$ to $0.299$.
This indicates that attribute-range and local-consistency constraints help optimize a cleaner explicit Gaussian representation.
The qualitative comparison in Figure~\ref{fig:qual_ablation_regularization} shows a more direct effect of this regularization.
Without Gaussian regularization, the model produces degenerate Gaussian attributes, leading to locally thin, transparent, and unstable semi-transparent regions.
This suggests that Gaussian regularization is not only beneficial for image metrics, but also necessary for preventing unstable Gaussian scales, opacities, and local inconsistencies during feed-forward prediction.

\paragraph{Geometry-guided Sampling and Implicit Decoding.}
\textit{w/o Geometry-guided Sampling} shows relatively moderate degradation in image metrics, but the full model still leads in PSNR, SSIM, and LPIPS on both ETH3D and ScanNet++.
For example, on ScanNet++, this variant drops from $22.240$ PSNR, $0.864$ SSIM, and $0.270$ LPIPS to $21.576$, $0.855$, and $0.274$, respectively.
This indicates that replacing geometry-guided sampled supports with pixel-aligned supports weakens the model, even when the subsequent implicit decoding is kept unchanged.
As shown in Figure~\ref{fig:qual_ablation_sampling_decoder}, the structural difference is more evident in visualization: \textit{w/o Geometry-guided Sampling} is more likely to create cracks along pixel-aligned supports under large side-view changes, suggesting that fixed pixel-aligned supports are less suitable for organizing Gaussians around slanted surfaces, object boundaries, and regions with depth variation.
\textit{w/o Implicit Decoder} degrades more clearly, reducing the ScanNet++ result to $20.811$ PSNR, $0.833$ SSIM, and $0.290$ LPIPS.
This shows that replacing query-conditioned decoding with dense pixel-aligned prediction weakens Gaussian attribute prediction.
Qualitatively, removing the implicit decoder amplifies the artifacts caused by fixed-grid prediction and produces more severe cracks, holes, and local surface breaks (refer to Figure~\ref{fig:qual_ablation_sampling_decoder}).
In contrast, the full model better preserves coherent surfaces and sharper object structures under large-baseline target views.

Overall, these quantitative and qualitative results show that the improvement comes from the joint effect of learned updates, dual-branch image conditioning, Gaussian regularization, geometry-guided sampling, and implicit decoding.

\paragraph{Support Budget.}
\begin{table}[t]
\centering
\setlength{\tabcolsep}{4.5pt}
\caption{Support budget ablation. "Inference" is representation inference time, and "Rendering" is Gaussian rendering time. Best results are highlighted in \protect\colorbox{green!30!yellow!30}{green}. The default setting uses $1.5\mathrm{M}$ sampled supports.}
\label{tab:support_budget}
\vspace{-1mm}
\begin{tabular}{lccccc}
\toprule
Budget & PSNR $\uparrow$ & SSIM $\uparrow$ & LPIPS $\downarrow$ & Inference $\downarrow$ & Rendering $\downarrow$ \\
\midrule
$0.5\mathrm{M}$ & 18.779 & 0.785 & 0.359 & \bestcell{0.444s} & \bestcell{0.004s} \\
$1.0\mathrm{M}$ & 21.180 & 0.848 & 0.294 & 0.552s & 0.007s \\
$1.5\mathrm{M}$ & \bestcell{22.240} & \bestcell{0.864} & \bestcell{0.270} & 0.598s & 0.009s \\
$2.0\mathrm{M}$ & 22.237 & \bestcell{0.864} & 0.271 & 0.772s & 0.011s \\
\bottomrule
\end{tabular}
\end{table}

Table~\ref{tab:support_budget} compares $0.5\mathrm{M}$, $1.0\mathrm{M}$, $1.5\mathrm{M}$, and $2.0\mathrm{M}$ supports on ScanNet++.
The results show that $0.5\mathrm{M}$ supports are clearly insufficient and cause a large quality drop.
After increasing the budget from $1.0\mathrm{M}$ to $2.0\mathrm{M}$, the metrics enter a diminishing-returns regime.
$2.0\mathrm{M}$ achieves nearly the same quality as $1.5\mathrm{M}$ but noticeably increases inference time.
Rendering time also grows with the support budget, from $0.004\mathrm{s}$ at $0.5\mathrm{M}$ supports to $0.011\mathrm{s}$ at $2.0\mathrm{M}$ supports.
Therefore, we use $1.5\mathrm{M}$ as the default support budget: it nearly reaches the quality plateau while keeping lower representation inference and rendering costs than $2.0\mathrm{M}$.

\paragraph{Prompt Depth Robustness.}
To assess the sensitivity of InfiniSplat-LiDAR to measurement perturbations in sparse depth conditioning, we conduct a controlled noise experiment on ETH3D. The results are reported in Table~\ref{tab:lidar_noise}.
Each source image is conditioned on 1,500 sparse depth prompts, and noise is applied only to the prompt depths as $d'=d(1+\epsilon)$, where $\epsilon\sim\mathcal{N}(0,\sigma^2)$.

\begin{table}[H]
\centering
\setlength{\tabcolsep}{7pt}
\caption{Robustness of InfiniSplat-LiDAR to multiplicative noise in sparse depth prompts on ETH3D. Best results are highlighted in \protect\colorbox{green!30!yellow!30}{green}.}
\label{tab:lidar_noise}
\vspace{-1mm}
\begin{tabular}{lccc}
\toprule
Noise $\sigma$ & PSNR $\uparrow$ & SSIM $\uparrow$ & LPIPS $\downarrow$ \\
\midrule
$0\%$ & \bestcell{25.880} & \bestcell{0.946} & \bestcell{0.151} \\
$1\%$ & 25.439 & 0.944 & 0.156 \\
$3\%$ & 24.337 & 0.937 & 0.170 \\
$5\%$ & 23.595 & 0.931 & 0.179 \\
\bottomrule
\end{tabular}
\end{table}

As the multiplicative noise level increases from $0\%$ to $5\%$, PSNR, SSIM, and LPIPS degrade smoothly and monotonically without abrupt failure.
This trend indicates that InfiniSplat-LiDAR degrades gracefully under small, controlled multiplicative perturbations to the sparse depth prompts.

\section{Limitations}
\begin{figure}[t]
    \centering
    \includegraphics[width=1.0\linewidth]{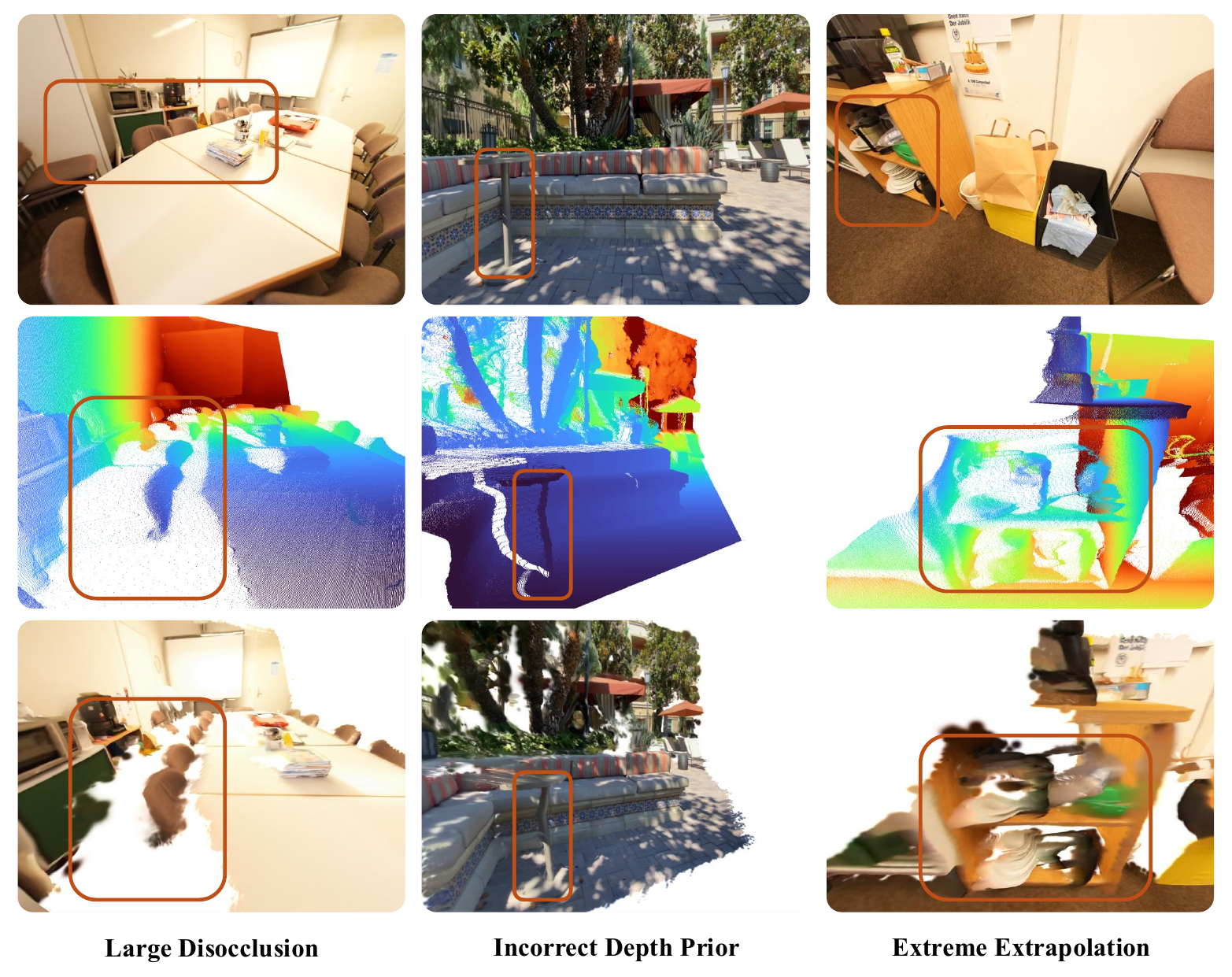}
    \caption{\textbf{Qualitative failure cases.}
    The columns show large disocclusion, an incorrect depth prior, and extreme extrapolation.
    From top to bottom, the rows show the input image, the depth predicted by the monocular depth model and warped to the selected target viewpoint, and the novel-view rendering of the Gaussian scene predicted by InfiniSplat.
    Brown boxes highlight the affected regions.
    The examples exhibit incomplete geometry under large disocclusion, artifacts inherited from incorrect depth priors, and severe structural distortion under extreme extrapolation.}
    \label{fig:failure_cases}
\end{figure}

Although InfiniSplat improves structural stability for feed-forward single-image 3DGS, it still faces geometric errors from both single-view ambiguity and imperfect pretrained depth priors.
Given only one input image, the model cannot observe occluded regions or disambiguate multiple plausible 3D explanations behind the visible pixels, so target views that expose large unseen regions, especially behind foreground objects or outside the source-view frustum, may produce incomplete geometry, stretched structures, or hallucinated appearance.
Moreover, geometry-guided support sampling relies on the predicted depth to construct the support distribution. When this prior fails on reflective surfaces, transparent objects, thin structures, textureless regions, or highly unusual scenes, the sampled supports may be placed on inaccurate geometric scaffolds.

Beyond these geometry-prior failures, InfiniSplat remains challenged by extreme viewpoint extrapolation and scenes whose structure or appearance cannot be captured well by a single depth-induced support distribution, such as non-Lambertian surfaces, repeated fine structures, very thin geometry, and strong depth discontinuities.
Representative failure cases are shown in Figure~\ref{fig:failure_cases}.
Future work could combine the framework with stronger generative priors, uncertainty-aware geometry estimation, or sparse multi-view inputs while retaining the efficiency of feed-forward Gaussian prediction.

\section{Conclusion}
We presented InfiniSplat, a feed-forward framework for single-image 3D Gaussian scene generation that combines geometry-guided support sampling with query-conditioned implicit Gaussian decoding.
By decoupling Gaussian prediction from fixed pixel centers and organizing supports around depth-induced surface structure, InfiniSplat moves single-image 3DGS toward a more surface-aligned representation for novel view synthesis.

Across cross-dataset evaluations on ETH3D, ScanNet++, Tanks and Temples, and DL3DV, InfiniSplat achieves state-of-the-art quantitative results against feed-forward baselines in both RGB-only and LiDAR-conditioned settings.
Qualitative comparisons and ablations further show that the proposed design produces fewer cracks and holes, cleaner normal organization, and more coherent surfaces under large viewpoint changes.
Taken together, these results suggest that feed-forward single-image 3DGS can move beyond pixel-aligned splat expansion toward a more structurally stable scene representation.

\bibliographystyle{ACM-Reference-Format}
\bibliography{reference}

\end{document}